%% file: main.tex
\documentclass[10pt,twocolumn,letterpaper]{article}

\usepackage{times}          
\usepackage{helvet}         
\usepackage{courier}        
\usepackage[margin=1in]{geometry} 
\usepackage{titlesec}
\titlespacing*{\section}{0pt}{*2}{*0.6}
\titlespacing*{\subsection}{0pt}{*1.6}{*0.4}
\titlespacing*{\subsubsection}{0pt}{*1.2}{*0.3}

\usepackage[utf8]{inputenc}
\usepackage{helvet}  
\usepackage{courier}  
\usepackage[hyphens]{url}  
\usepackage{graphicx} 
\usepackage{natbib}  
\usepackage[table]{xcolor}
\usepackage{enumitem}

\usepackage{diagbox}
\usepackage{algorithm}
\usepackage{algorithmic}
\usepackage{colortbl}
\usepackage{hyperref}

\hypersetup{
    colorlinks,
    linkcolor={blue!80!black},
    citecolor={green!60!black},
    urlcolor={red!60!black},
    linkbordercolor={blue!50!white}, 
    citebordercolor={green!50!white},
    urlbordercolor={red!50!white},
    pdfborderstyle={/S/U/W 1}, 
}

\author{\and \hspace{-2em}Luca-Andrei Fechete\thanks{École Polytechnique, Palaiseau, France (Research Intern)}~\protect\thanks{Paris Research Center, Huawei Technologies, Boulogne-Billancourt, France}\and
Mohamed Sana\footnotemark[2] \and
Fadhel Ayed\footnotemark[2] \and
Nicola Piovesan\footnotemark[2] \and 
Wenjie Li\footnotemark[2] \and
Antonio De Domenico\footnotemark[2] \and 
Tareq~Si~Salem\footnotemark[2]~\thanks{Lead researcher for this study.  Corresponding author: \href{mailto:tareq.si.salem@huawei.com}{tareq.si.salem@huawei.com}}
}
\definecolor{Teal}{rgb}{0.0, 0.5, 0.5}
\usepackage{newfloat}
\usepackage{listings}
\usepackage{rotating}
\usepackage{nicefrac}  
\usepackage{amsfonts}
\usepackage{amssymb,amsmath,amsthm}
 \usepackage{multirow} 
\usepackage{subcaption}

\usepackage{booktabs}
\input{notation_defs}

\title{\textbf{Goal-Oriented Time-Series Forecasting:\\Foundation Framework Design}}

\date{}
\begin{document}
\maketitle
\begin{abstract}
Conventional time-series forecasting methods typically aim to minimize overall prediction error, without accounting for the varying importance of different forecast ranges in downstream applications. We propose a training methodology that enables forecasting models to adapt their focus to application-specific regions of interest at inference time, without retraining. The approach partitions the prediction space into fine-grained segments during training, which are dynamically reweighted and aggregated to emphasize the target range specified by the application. Unlike prior methods that predefine these ranges, our framework supports flexible, on-demand adjustments. Experiments on standard benchmarks and a newly collected wireless communication dataset demonstrate that our method not only improves forecast accuracy within regions of interest but also yields measurable gains in downstream task performance. These results highlight the potential for closer integration between predictive modeling and decision-making in real-world systems.
\end{abstract}

\section{Introduction}
Time-series forecasting (TSF) represents a significant area of study within machine learning (ML), with practical applications demonstrable in various domains, including but not limited to, economics~\cite{granger2014forecasting}, energy resource management~\cite{martin2010prediction, qian2019review}, transportation optimization~\cite{chen2001freeway}, meteorological prediction~\cite{wu2023interpretable}, inventory optimization~\cite{li2022demand}, and healthcare~\cite{ghassemi2015multivariate,ray2025flusion}.  At its core, TSF is concerned with constructing predictive models for time-dependent sequential data. This involves leveraging historical patterns and relationships within observations to forecast future data points.  The methodologies employed in TSF are diverse, ranging from classical statistical approaches, such as Autoregressive Integrated Moving Average (ARIMA)~\cite{hyndman20158} and Exponential Smoothing (ETS)~\cite{brown1956exponential}, to deep learning (DL) approaches, including Multi-Layer Perceptrons (MLPs)~\cite{zeng2023transformers}, Recurrent Neural Networks (RNNs)~\cite{zhang1998time}, Long Short-Term Memory (LSTMs)~\cite{siami2019performance}, Temporal Convolutional Networks (TCNs)~\cite{he2019temporal}, and  Transformer architectures~\cite{nie2022time, liu2023itransformer}. More recently, the field has witnessed the appearance of foundation Large Time-Series Models (LTSMs), pre-trained on extensive time-series datasets to enable zero-shot forecasting. These models, including Timer~\cite{liu2024timer}, Moirai~\cite{woo2024unified}, TimesFM~\cite{das2024decoder}, Chronos~\cite{ansari2024chronos}, Moment~\cite{goswami2024moment}, and Toto~\cite{cohen2024toto} predominantly utilize variations of the Transformer architecture.

Generally, TSF methodologies prioritize the minimization of predictive error, often neglecting the integration of predicted outputs within subsequent downstream  processes. In numerous downstream applications of forecasting, the practical significance of forecast errors is not uniform, and this treatment introduces suboptimal model performance with respect to the ultimately desired objective. This is effectively illustrated by forecasting challenges \emph{IEEE-CIS Technical Challenge on Predict+Optimize for Renewable Energy Scheduling}~\cite{bergmeir2022comparison} and the \emph{M5 Accuracy Competition}~\cite{bergmeir2022comparison}, where evaluations based solely on forecast accuracy substantially mismatches with evaluations based on the eventual optimization objective, such as energy minimization. This necessitates the development of TSF models that explicitly incorporate downstream task objectives during development and evaluation. This integration is essential given the widespread of real world analytical systems combining predictive and optimization components. For example, Bertsimas and Kallus~\cite{bertsimas2020predictive} propose learning weight functions from data for integration into optimization objectives. Similarly, Elmachtoub and Grigas~\cite{elmachtoub2022smart} proposed a ``Predict-then-Optimize'' framework within linear programming, directly leveraging optimization structure to inform loss function design. Furthermore, the ML community has witnessed a significant trend towards end-to-end (E2E) learning paradigms, applied across domains such as finance~\cite{bengio1997using}, image recognition~\cite{wang2011end}, robotic manipulation~\cite{levine2016end}, and inventory-management~\cite{qi2023practical} highlighting the potential for this approach in TSF.

\begin{figure}[t!]
 \centering
 \includegraphics[width=1\linewidth]{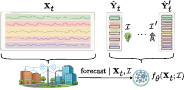}
 \caption{The figure illustrates time-series forecasting in a wireless network, highlighting the stages of data collection, model training, and application to downstream tasks such as energy efficiency---where accurate prediction of low-traffic periods is critical---and power allocation, which requires precise forecasts across two traffic bands. While traditional forecasting approaches overlook such task-specific requirements, the proposed method enables a single model to adapt to diverse downstream tasks at inference time, offering improved flexibility over task-specific end-to-end systems.}
 \label{fig:illustration-E2E-Found}
\end{figure}

Existing TSF methods~\cite{bertsimas2020predictive, elmachtoub2022smart, qi2023practical} typically assume fixed, pre-defined task specifications, with regions of forecasting importance known and static. In practice, many applications, such as wireless traffic prediction \cite{Wu2021DeepBSC, traffic_1}, require adaptation to unknown and dynamically changing importance regions. For example, energy efficiency policies may prioritize low-traffic periods for base-station deactivation, while power allocation strategies demand sensitivity to both low and high extremes. These thresholds are rarely available a priori and may vary over time, motivating TSF frameworks that allow post-hoc configuration of importance during inference. An illustrative example is shown in Figure~\ref{fig:illustration-E2E-Found}.
\paragraph{Contributions.} In this work, we address this gap by: (1)~proposing a training methodology that extends multivariate TSF models to adapt to multiple downstream tasks at inference without retraining, (2)~conducting extensive experiments on synthetic and real-world traces, including comprehensive baseline comparisons and ablation studies, to validate its effectiveness, and (3)~introducing a new wireless mobile network dataset to support further research. 

\paragraph{Outline of Paper.}  The remainder of this paper is organized as follows. Section~\ref{s:related_work} reviews relevant literature. Section~\ref{sec:tsfproblem} formalizes the forecasting problem. Section~\ref{sec:methodology} presents the proposed methodology. Section~\ref{sec:experiements} describes the experimental setup and evaluates the performance of the proposed training approach. Finally, Section~\ref{s:conclusion} concludes the paper and discusses directions for future work.
\section{Literature Review}

\paragraph{Time-Series Forecasting.} \label{s:related_work}
Traditionally, TSF has been dominated by statistical models such as ARIMA~\cite{hyndman20158} and ETS~\cite{brown1956exponential}. With the increase in computational resources and the availability of large-scale datasets, deep learning methods have become increasingly prevalent, including MLPs~\cite{zeng2023transformers}, RNNs~\cite{zhang1998time}, LSTMs~\cite{siami2019performance}, TCNs~\cite{he2019temporal}, and, more recently, Transformer-based architectures~\cite{nie2022time, liu2023itransformer, wu2021autoformer, zhou2021informer, zhou2022fedformer}. Research on adapting Transformers for TSF has focused on three main directions: refining internal components such as attention mechanisms~\cite{wu2021autoformer, zhou2021informer, zhou2022fedformer}; improving input token representations through approaches like stationarization~\cite{liu2022non}, channel independence~\cite{nie2022time}, and patching~\cite{nie2022time, liu2023itransformer}; and modifying the overall Transformer architecture and its modules~\cite{zhang2023crossformer}. More recently, foundation LTSMs such as Timer~\cite{liu2024timer}, Moirai~\cite{woo2024unified}, TimesFM~\cite{das2024decoder}, Chronos~\cite{ansari2024chronos}, Moment~\cite{goswami2024moment}, and Toto~\cite{cohen2024toto} have emerged, leveraging large-scale pretraining and Transformer variants to enable zero-shot forecasting. Building on these advances, we propose an extension of state-of-the-art TSF architectures that incorporates a task-specific configuration mechanism, allowing dynamic adaptation of predictions during inference.
\paragraph{Task-Aware Forecasting.}
Prior work on task-aware TSF has explored modifying the training objective through loss reshaping or reweighting~\cite{park2023deep, xue2023card, cheng2023fitting, hounie2024loss}, typically to integrate uncertainty quantification or fairness considerations rather than to optimize for specific downstream tasks. Another line of research has focused on frameworks that derive loss weights from downstream task characteristics to guide model training~\cite{bergmeir2022comparison, elmachtoub2022smart, grabocka2019learning}. While our approach is conceptually related to this latter direction through its use of loss shaping, it differs in its ability to train a single model that can flexibly adapt to a wide range of downstream task requirements at inference time.

\section{The Forecasting Problem} \label{sec:tsfproblem}
\paragraph{Time-Series.} A multivariate time-series is a sequence of vector-valued observations ordered temporally. Let $\vec{x}_t \in \mathbb{R}^n$ denote a $n$-dimensional real-valued vector observed at time $t$, where $t \in [T]$ and $n \geq 1$. The full time-series is represented as the ordered set $\set{\vec{x}_1, \vec{x}_2, \ldots, \vec{x}_T}$, where each $\vec{x}_t = \parentheses{x_{t,i}}_{i \in [n]}$ corresponds to measurements of $n$ variables at time $t$. The temporal dependency between observations is intrinsic to the series, with $\vec{x}_t$ potentially influenced by past values $\vec{x}_{t-k}$ for some lag $k > 0$. We assume each series $i \in [n]$ lies within a bounded set  $\mathcal{X} \subset \reals$, where $\card{\mathcal X} < \infty$.

\paragraph{The Learning Problem.} Time-series forecasting is reformulated as a supervised regression task by constructing input-output pairs from sliding windows over the series. Let $\tau \in \naturals$ be the forecast horizon and $w\in \naturals$ be the window size. In practice, the window size $w$ is treated as a hyperparameter. For each time $t$, the input $\vec X_t = \vec{x}_{t-w:t-1} \in \mathcal{X}^{w \times n}$ consists of a history window $\{\vec{x}_{t-w}, \ldots, \vec{x}_{t-1}\}$, and the target $\vec{Y}_t = \vec {x}_{t:t+\tau-1} \in \mathcal{X}^{\tau \times n}$ is the future window $\{\vec{x}_{t}, \ldots, \vec{x}_{t+\tau-1}\}$. 
We consider a parametric mapping
\begin{align}
f_{\vec{\theta}} &: \mathcal{X}^{w \times n} \to \mathcal{X}^{\tau \times n},
\end{align}
trained to approximate the unknown dynamical operator 
\begin{align}
    f &: \mathcal{X}^{w \times n} \to \mathcal{X}^{\tau \times n},
\end{align}
The observations satisfy
\begin{align}
\vec{Y}_t &= f(\vec{X}_t) + \boldsymbol{\epsilon}_t, \quad \text{for}~t \in [T], \label{eq:model}
\end{align}
where $\boldsymbol{\epsilon}_t \in \mathbb{R}^{\tau \times n}$ is zero-mean noise with finite covariance.  This generates a dataset $D = \{(\vec{X}_t, \vec{Y}_t)\}_{t=w}^{T-\tau}$. The goal is to learn the model $\vec \theta_\star \in \reals^d$ for $d\geq 1$ that minimizes the expected loss over the data distribution $\mathcal{D}(\vec{X}, \vec{Y})$:
\begin{align}
 \vec \theta_\star \in \argmin_{\vec \theta \in \reals^d} \mathbb{E}_{(\vec{X}, \vec{Y}) \sim \mathcal{D}} \interval{ l(f_{\vec \theta}(\vec{X}), \vec{Y}) },
\end{align}
where $l: \mathbb{R}^{\tau \times n} \times \mathbb{R}^{\tau \times n} \rightarrow \mathbb{R}$ is a differentiable loss function. In practice, this expectation is approximated by the empirical risk over $D$: ${\vec{\tilde{\theta}}}_\star \in \argmin_{\theta} \frac{1}{|D|} \sum_{(\vec{X}_t, \vec{Y}_t) \in D} l (f_{\vec \theta}(\vec{X}_t), \vec{Y}_t) + R(\vec \theta )$. The regularization term $R: \reals^d \to \reals$ (e.g., the Euclidean distance $R(\vec \theta) = \lambda \norm{\vec \theta}^2_2$) is incorporated to control the complexity of the model parameters $\vec \theta$,  and it aims to prevent the memorization of noise within the training data,  enhancing the model's ability to generalize to novel data~\cite{shalev2014understanding}. 
Among the most commonly used loss functions are the {Mean Squared Error (MSE)} and the {Mean Absolute Error (MAE)}, each of which serves a distinct statistical purpose (i.e., conditional expectation and the conditional median, respectively). 


\section{Methodology} \label{sec:methodology}
Our methodological investigation proceeds by first establishing a baseline policy (\baselinePolicy{}), representative of standard forecasting model training that overlooks specific prediction intervals. We then define an end-to-end policy (\taskSpecificPolicy{}), which focuses on training a model tailored to a particular interval of interest. Subsequently, we introduce a naive policy (\uniformPolicy{}) designed to train a foundation model capable of adapting to various intervals at inference time by exposing it to all possible intervals during training. Following this, we describe a policy (\discretePolicy{}) that explores a finite subset of relevant intervals, strategically chosen such that their combination covers the entire forecasting space. Finally, we propose a policy (\discretePacthedPolicy{}) that discretizes the forecasting space and, when integrated with a patching technique, can enable the training of more effective foundation models.
\paragraph{Baseline Policy (\baselinePolicy{}).} This policy follows the  baseline training procedure detailed in Section~\ref{sec:tsfproblem}. This training approach does not incorporate interval sensitivity, and the loss is computed with respect {to} the forecasting target over  the domain $\mathcal X$. 
\paragraph{Task-specific Policy (\taskSpecificPolicy{}).} Given a target interval $\mathcal{I} \subseteq \mathcal{X}$, the task-specific policy only considers a forecasting target within this range. In particular, this policy can be considered as a variation of the baseline policy  limited to consider a forecasting loss  solely over the interval{s} of interest. Our goal is to learn the model $\vec \theta_\star \in \reals^d$  that minimizes the interval-specific expected loss over a data distribution $\mathcal{D}(\vec{X}, \vec{Y})$  given as 
\begin{align}
  {\vec\theta}_\star \in \argmin_{\vec \theta \in \reals^d} 
\underset{(\vec{X}, \vec{Y}) \sim \mathcal{D}}{\mathbb{E}} 
&\Big[ l\parentheses{f_{\vec \theta}(\vec{X}) , \vec{Y}}\nonumber \\ &\cdot \mathds{1} \parentheses{\vec{Y} \in \mathcal{I}^{\tau \times n}}\Big], 
\end{align}
where $\mathds{1} (\chi) \in \set{0,1}$ is the indicator function  equal to 1 when condition $\chi$ is true. Again, the expectation can be approximated  by the interval-specific empirical risk over some dataset $D$ consisting of samples from $\mathcal{D}$ with a regularization term.
\paragraph{Continuous-Interval Training Policy (\uniformPolicy{}).} To allow the forecasting model to differentiate between any target intervals within $\mathcal{X}$, we incorporate the interval as a covariate. A covariate interval $\mathcal{I} \subseteq \mathcal{X}$ is represented as a two-dimensional vector in $\mathcal{X}^2$ containing its boundary values {(e.g., a vector encoding $\parentheses{I_{\min}, I_{\max}} \in \mathcal{X}^2$, for some interval $[I_{\min}, I_{\max}]\subseteq\mathcal{X}$)}.  This enables the model to learn the relationship between varying intervals and its predictive outputs.  Specifically, the learned mapping $f_{\vec \theta}$ becomes a function of both the input time-series $\vec X$ and the interval $\mathcal{I}$, i.e., the mapping $f_{\vec \theta}: \mathcal{X}^{w \times n + 2}  \rightarrow \mathcal{X}^{\tau \times n}$. During training, we sample intervals uniformly at random over the set $\mathcal{U}_\delta = \set{\mathcal{I} \subset \mathcal{X}: \card{\mathcal{I}} \geq \delta} $, where $\delta \in [0, |\mathcal{X}|)$ represents the minimum length of the sampled intervals. The minimal distance constraint is introduced to make the training more stable, because small intervals contain less samples introducing high variance.  Consequently, for a data sample $(\vec{X}_t, \vec{Y}_t) \sim \mathcal D$ and its corresponding sampled interval $\mathcal{I}_t$, our custom loss function is defined as:
\begin{align}
 \loss(f_{\vec \theta}(\vec{X}_t, \mathcal{I}_t) , \vec{Y}_t, \mathcal{I}_t) = &l\parentheses{f_{\vec \theta}(\vec{X}_t, \mathcal{I}_t), \vec{Y}_t}\nonumber\\
 &\cdot \mathds{1}(\vec{Y}_t \in \mathcal{I}_t^{\tau\times n}).\label{eq:hard_target}
\end{align}
Thus, our training objective is to learn the model parameters $\vec \theta^\star\in \mathbb{R}^d$ ($d \geq 1$) that minimize the interval-uniform expected loss over the data distribution $\mathcal{D}(\vec{X}, \vec{Y})$ and the interval distribution $\mathcal{U}_\delta$:
\begin{align}
 \vec \theta_\star \in \argmin_{\vec \theta \in \mathbb{R}^d}  \mathbb{E}_{(\vec{X}, \vec{Y}) \sim \mathcal{D}, \mathcal{I} \sim \mathcal{U}_\delta} \interval{ \loss(f_{\vec \theta}(\vec{X}, \mathcal{I}) , \vec{Y}, \mathcal{I})}.
\end{align}
Similar to the previous policies, the above problem is approximated by minimizing the empirical risk over the dataset $D$ and the interval distribution $\mathcal{U}_\delta$: $ \tilde{\vec\theta}_\star  \in \argmin_{\vec \theta \in \mathbb{R}^d} \frac{1}{|D|} \sum_{(\vec{X}_t, \vec{Y}_t) \in D} \loss(f_{\vec \theta}(\vec{X}_t, \mathcal{I}_t) , \vec{Y}_t, \mathcal{I}_t) + R(\vec \theta )$,  where $\mathcal{I}_t$ is the sampled interval from $\mathcal{U}_\delta$ for each $(\vec{X}_t, \vec{Y}_t) \in D$.
\paragraph{Discretized-Interval Training Policy  (\discretePolicy{}).}
This policy closely resembles the \uniformPolicy{}, except that the support space of the distribution of possible interval is severely reduced. In particular, the interval selection process now involves sampling  an interval from a discrete distribution, denoted by $\mathcal{C}_L$, defined over a set of $L$ disjoint intervals that collectively cover the interval  $\mathcal{X}$, i.e., for a given $L$ the support $\supp{\mathcal{C}_L}$ satisfies  $\dot{\bigcup}_{\mathcal{I} \in \supp{\mathcal{C}_L}}~\mathcal{I} = \mathcal X$, where $\dot{\bigcup}$ denotes the union of disjoint sets.
\begin{figure}[t!]
    \centering
    \includegraphics[width=0.8\linewidth]{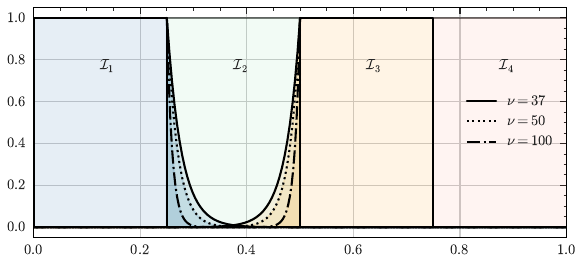}
    \caption{Impact of decay on the overlap of four intervals spanning the range $[0,1]$. The figure illustrates the weighting associated with the intervals $\mathcal{I}_1 = [0, 0.25]$ and $\mathcal{I}_3$ for various decay rates $\nu \in \set{37, 50, 100}$. Note how increasing the decay rate reduces the overlap between the intervals. For $\nu = 37$, the weight value reaches $1\%$ at the midpoint of the adjacent interval.}
    \label{fig:decay_function}
\end{figure}
\paragraph{Patching-Augmented Discretized Training Policy (\discretePacthedPolicy{}).} The premise behind the design of this policy is to consider the \discretePolicy{} configured with a sufficiently large $L$ corresponding to a finer discretization of the domain $\mathcal{X}$. At inference time when requested with an unknown interval $\mathcal{I}$, the policy combines the constituent intervals to make the predictions. This is made possible with two main modifications. First, the loss function incorporates a decay function that modulates the contribution of each sample to the loss based on its distance from the center of its associated interval.\footnote{This function is depicted  in Figure~\ref{fig:decay_function}.} This decay mechanism is defined as 
\begin{align}
  d_\nu(y, \mathcal{I}) \triangleq \exp\left(-\nu \max\left(0, \left| y - \Delta_\mathrm{avg} \right| - \Delta_\mathrm{diff}\right)\right),\label{eq:soft_target}
\end{align}
 where $\nu$ is the decay rate, and $\Delta_\mathrm{avg}\triangleq \parentheses{\max(\mathcal{I}) + \min(\mathcal{I})}/{2}$ is the midpoint of an interval and $\Delta_\mathrm{diff} \triangleq \parentheses{\max(\mathcal{I}) - \min(\mathcal{I})}/{2}$ is half of the length of the interval. 
 
As the decay rate $\nu$ becomes large ($\nu \to \infty$), the weight $\prod_{i \in [n], t \in [\tau]}d_\nu(y_{i}, \mathcal{I}) \in [0,1]$ converges to the indicator function $\mathds{1}(\vec Y \in \mathcal{I}^{n \times \tau}) \in \{0,1\}$ in Eq.~\eqref{eq:hard_target}. This weighting allows learning at interval boundaries and introduces a soft overlap between intervals, which is beneficial for reducing patching errors caused by boundary uncertainties. Additionally, a classification head $f_{\vec \theta}^c : \mathcal{X}^{w \times n + 2} \to [0,1]^{\tau \times n}$ (parameterized by a shared $\vec{\theta}$) is appended to predict if the true value falls within $\mathcal{I}_t$, using the same notation as before. The modified regression loss with the soft boundary is 
\begin{align}
    &\loss_\nu(f_{\vec \theta}(\vec{X}_t, \mathcal{I}_t) , \vec{Y}_t, \mathcal{I}_t) \\ &\triangleq l(f_{\vec \theta}(\vec{X}_t, \mathcal{I}_t), \vec{Y}_t)\nonumber
     \!\!\!\!\!\!\!\prod_{i \in [n], t \in [\tau]}\!\!\!d_\nu(y_{t,i}, \mathcal{I}_t), 
\end{align}
and the modified classification loss is \begin{align}
    &{\loss_\nu^\prime}(f_{\vec{\theta}}^c(\vec{X}_t, \mathcal{I}_t) , \mathds{1}(\vec{Y}_t \in \mathcal{I}_t), \mathcal{I}_t) \nonumber\\
    &\triangleq {l_\nu^\prime} (f_{\vec{\theta}}^c(\vec{X}_t, \mathcal{I}_t) , \mathds{1}(\vec{Y}_t \in \mathcal{I}_t)) \!\!\!\!\!\!\!\prod_{i \in [n], t \in [\tau]}\!\!\!d_\nu(y_{t,i}, \mathcal{I}_t),
\end{align} 
where $l_\nu^\prime$ is a classification loss like cross entropy. The learning objective for the augmented model parameters (including the classifier) is defined as follows:
\begin{align}
{\vec\theta}_\star \in \argmin_{\vec{\theta} \in \mathbb{R}^d}\,\,&\mathbb{E}_{(\vec{X}, \vec{Y}) \sim \mathcal{D}, \mathcal{I} \sim \mathcal{C}_L} \big[\loss_\nu(f_{\vec{\theta}}(\vec{X}, \mathcal{I}) , \vec{Y}, \mathcal{I}) \nonumber\\
&+ \phi \cdot { \loss_\nu^\prime}(f_{\vec{\theta}}^c(\vec{X}, \mathcal{I}) , \mathds{1}(\vec{Y} \in \mathcal{I}), \mathcal{I})\big],
\end{align}
where $\phi \in [0,1]$ is a hyperparameter introduced to balance the importance of the regression and classification tasks.
This objective is approximated empirically using the dataset $D$ and the interval distribution $\mathcal{C}_L$:
\begin{align}
\tilde{\vec\theta}_\star \!\! \in\! \,\,&\argmin_{\vec{\theta} \in \mathbb{R}^d}\frac{1}{|D|}  \!\sum_{(\vec{X}_t, \vec{Y}_t) \in D}\!\! \big( \loss_\nu(f_{\vec{\theta}}(\vec{X}_t, \mathcal{I}_t) , \vec{Y}_t, \mathcal{I}_t)\nonumber \\&+\! \phi \cdot {\loss_\nu^\prime}(f_{\vec{\theta}}^c(\vec{X}_t, \mathcal{I}_t) , \mathds{1}(\vec{Y}_t \in \mathcal{I}_t), \mathcal{I}_t) \big) \!+\! R(\vec \theta),
\end{align}
where $\mathcal{I}_t$ is the sampled interval from $\mathcal{C}_t$ for each $(\vec{X}_t, \vec{Y}_t) \in D$.

\noindent \emph{Patching Mechanism.} Given an arbitrary interval $\mathcal{I}$, we first identify the subset of training intervals that intersect with ${\mathcal{I}}$, given by the mapping  
\begin{align}
    \Xi_L (\mathcal{I}) \triangleq \set{ \mathcal{I}' \in \supp{\mathcal{C}_L} : \mathcal{I}' \cap {\mathcal{I}} \neq \emptyset },
\end{align}
where $\supp{\mathcal{C}_L}$ is the support of the distribution $\mathcal{C}_L$ of size $L$. The logic behind this is that when $L$ is large enough, the union of intervals in $\Xi_L(\mathcal{I})$ is a good approximation of $\mathcal{I}$. For an input series $\vec X$ and target interval $\mathcal{I}$ the patching mechanism strategies are defined as follows: 
\begin{itemize}
    \item \emph{Averaging Strategy ({1-strategy}).} This strategy computes a weighted average of the regression predictions for all intersecting training intervals, weighted by their respective classification probabilities for the given input and the target interval $\mathcal{I}$. Formally, defined as
    \begin{align}
         \hat{\vec{Y}} \triangleq \frac{\sum_{\mathcal{I}^\prime \in \Xi_L(\mathcal{I})} f_\theta^c(\vec{X}, \mathcal{I}^\prime) f_\theta(\vec{X}, \mathcal{I}^\prime)}{\sum_{\mathcal{I}^\prime \in \Xi_L(\mathcal{I})} f_\theta^c(\vec{X}, \mathcal{I}^\prime)}.
    \end{align}
    \item \emph{Maximum Confidence Strategy ({$\infty$-strategy}).} This strategy selects the regression prediction corresponding to the training interval with the highest classification probability for the given input and the target interval $\mathcal{I}$. Formally, defined as
    \begin{align}
        \hat{\vec{Y}} \triangleq f_\theta \big({\vec{X}, \underset{\mathcal{I}^\prime \in \Xi_L(\mathcal{I})}{\argmax}~ f_\theta^c(\vec{X}, \mathcal{I}^\prime) }\big).
    \end{align}
\end{itemize}

\section{Experiments} \label{sec:experiements}
This section reports the numerical results obtained from the proposed training policies and evaluates their forecasting performance in comparison with multiple baseline methods. We begin by describing the forecasting models, training policies, benchmark datasets, and training configurations employed. This is followed by both qualitative and quantitative analyses of the various training strategies applied to these models.\footnote{The code for all experiments is available at  \href{https://github.com/netop-team/gotsf}{Reproducibility Code}.}
\subsection{Experimental Setup}
\paragraph{Time-Series Forecasting Models.} We evaluate four state-of-the-art TSF models: iTransformer~\cite{liu2024itransformerinvertedtransformerseffective}, DLinear~\cite{zeng2023transformers}, PatchTST~\cite{nie2023timeseriesworth64}, and TimeMixer~\cite{wang2024timemixerdecomposablemultiscalemixing}. To incorporate target interval information $\mathcal{I}$, we modify each architecture accordingly. Given an interval of interest $\mathcal{I} \subseteq \mathcal{X}$, the vectorized form of $\mathcal{I}$ is concatenated with the temporal encoding in iTransformer, while for DLinear, PatchTST, and TimeMixer, it is introduced as two additional temporal channels. We further extend these regression models to support a dual-task objective---forecasting and classification---by adding a classification head. This is implemented by doubling the output dimension of the final projection layer (equal to the forecasting horizon $\tau$) and partitioning it into regression outputs (first $\tau$ dimensions) and classification logits (remaining $\tau$ dimensions). This enables the models to both predict future values and estimate the probability that these forecasts fall within the target interval $\mathcal{I}$.

\paragraph{Training Policies.} We evaluate the training policies introduced in Section~\ref{sec:methodology}, namely \baselinePolicy{}, \taskSpecificPolicy{}, \uniformPolicy{}, \discretePolicy{}, and \discretePacthedPolicy{}. In the latter, the $\star$ symbol indicates the patching method used, with $\star = 1$ denoting average patching and $\star = \infty$ denoting maximum confidence patching.

\begin{figure}[t!]
 \centering
 \subcaptionbox{Noise-free hypotheses in \synthds{}\label{dif:synthds}}{\includegraphics[width=0.48\linewidth]{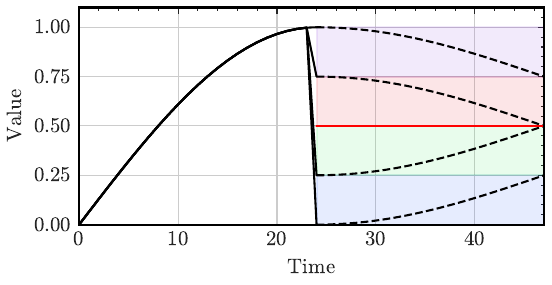}}
 \subcaptionbox{Averaging tendencies of \baselinePolicy{} \label{fig:avgtend}}{\includegraphics[width=0.47\linewidth]{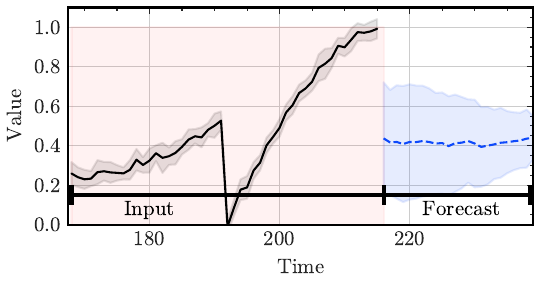}}
 \caption{ Subfigure (a) depicts the expected hypotheses used to construct the synthetic trace \synthds{}. Subfigure~(b) depicts the inability of the baseline policy \baselinePolicy{} to distinguish between the different patterns in \synthds{} dataset, as it predicts the average hypothesis (red line at 0.5). The model used is a trained iTransformer. Each subplot shows a time-shift of six steps. The black dotted line indicates the true values. The blue line shows the model's predictions.  over 10 random seeds.}
\end{figure}

\paragraph{Benchmarking Datasets.}  In our experimental evaluation, we employ several benchmark datasets to substantiate our claims:  
\begin{enumerate}
    \item \textbf{\synthds{}} is a synthetic dataset designed to highlight the various components of our proposed methodology. The trace is constructed by combining an input signal  
    \begin{align}
        \textstyle   \parentheses{\sin\parentheses{\frac{\pi n}{2D}}}_{n \in [w]},
    \end{align}
    where $w = 24$ denotes the signal length, with an output signal selected uniformly at random (u.a.r.) from the set  
    \begin{align}
        \textstyle \set{ \frac{1}{4}\parentheses{ \sin\parentheses{\frac{\pi}{2w}n + \frac{\pi}{2}} + k}_{n \in [w]} : k \in [4] }.
    \end{align}
    To produce a trace spanning $T = 3.1 \times 10^3$ timesteps, the same randomly selected signal is concatenated multiple times. Gaussian noise with mean and standard deviation $0.05$ is then added to the trace. A noise-free version, without the addition of Gaussian noise, is shown in Figure~\ref{dif:synthds}.
    \item \textbf{\wirelessds{}}~\cite{huggingfaceNetopBeamLevelTrafficTimeseriesDatasetDatasets} is released to the public domain as part of this study.\footnote{The dataset is publicly available at \href{https://huggingface.co/datasets/netop/Beam-Level-Traffic-Timeseries-Dataset}{Beam-Level Time-Series Dataset}.} We consider a subset of the DLPRB modality, which contains wireless beam-level  measurements for $100$ beams over $10^3$ timesteps.
    \item \textbf{\texttt{Traffic}}~\cite{TrafficDataset} contains hourly road occupancy rates (ranging from 0 to 1) collected by sensors on San Francisco Bay Area freeways between 2015 and 2016, covering a total of 48 months.
    \item \textbf{\texttt{Electricity}}~\cite{ElectricityDataset} records hourly electricity consumption (in kWh) for 321 clients from 2012 to 2014.
    \item \textbf{\texttt{Weather}}~\cite{wu2021autoformer} contains 21 meteorological variables, including air temperature and humidity, recorded every 10 minutes throughout 2020.
\end{enumerate}

\begin{figure}[t!]
    \centering
        \subcaptionbox{\label{fig:e2e_unif_a}}{\includegraphics[width=0.47\linewidth]{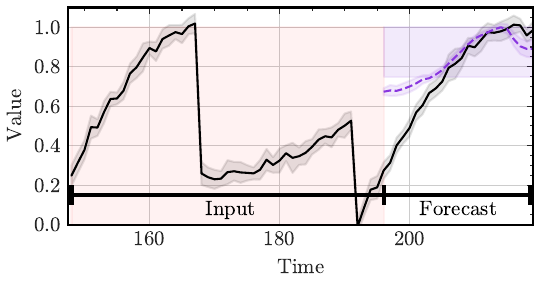}}
             \subcaptionbox{\label{fig:e2e_unif_b}}
                 {\includegraphics[width=0.47\linewidth]{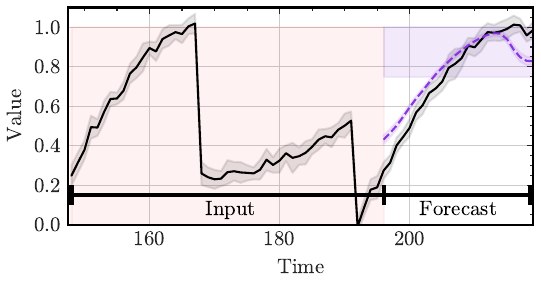}}
            \subcaptionbox{\label{fig:disc_patch_4}}{\includegraphics[width=0.47\linewidth]{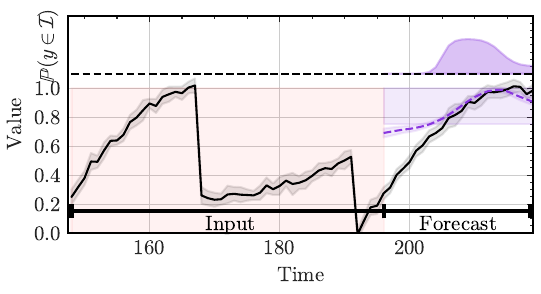}}
                 \subcaptionbox{\label{fig:disc_patch_8}}{\includegraphics[width=0.47\linewidth]{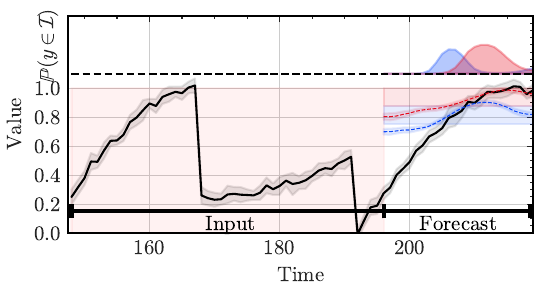}}
                \subcaptionbox{\label{fig:disc_patch_8_1}}{\includegraphics[width=0.47\linewidth]{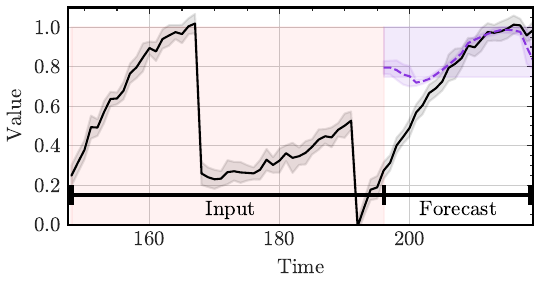}}
                  \subcaptionbox{\label{fig:disc_base}}{\includegraphics[width=0.47\linewidth]{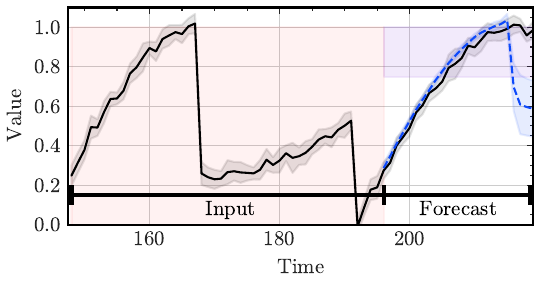}}
                  
    \caption{Subfigures~(a)–(f) present the performance of the iTransformer model trained on the \synthds{} trace. Each subfigure corresponds to a different policy: \texttt{E2E}, \texttt{C}, \texttt{D$_4$}, \texttt{D$_8$}, \texttt{D$_8^1$}, and \texttt{B}, respectively. The purple region marks the interval of interest, $\mathcal{I} = [0.75, 1.0]$. In the $1$-strategy, eight intervals are patched into four. The filled region represents the classifier’s predicted probability $\mathbb{P}(y \in \mathcal{I})$. Under the \texttt{D$^1_8$} policy, two intervals must be patched to generate a forecast for the selected interval $\mathcal{I}$. Black lines indicate target forecast values, while dashed lines denote the mean prediction over time, with shaded areas showing the standard deviation computed over $10$ random seeds.
}
    \label{fig:placeholder}
\end{figure}

\begin{table*}[t!]
\centering
\setlength{\tabcolsep}{3pt}
\renewcommand{\arraystretch}{.7}
\begin{adjustbox}{width=\linewidth}
\begin{tabular}{||c||c|c|c|c|c|c||c|c|c|c|c|c||c|c|c|c|c|c||c|c|c|c|c|c||}
\toprule
\textbf{Models} & \multicolumn{6}{c|}{\textbf{DLinear}} &
   \multicolumn{6}{c|}{\textbf{TimeMixer}} &
   \multicolumn{6}{c|}{\textbf{PatchTST}} &
   \multicolumn{6}{c||}{\textbf{iTransformer}}\\
\midrule
\diagbox[width=5em]{\scriptsize\textbf{Intervals}}{\scriptsize\textbf{Policies}} &
  \texttt{D}$_L$ & \texttt{D}$_{2L}^1$ & \texttt{D}$_{2L}^\infty$ & \texttt{B} & \texttt{C}$_{0.2}$ & \shortstack{\textbf{\textcolor{teal}{Impro-}}\\\textbf{\textcolor{teal}{vement}}} &
  \texttt{D}$_L$ & \texttt{D}$_{2L}^1$ & \texttt{D}$_{2L}^\infty$ & \texttt{B} & \texttt{C}$_{0.2}$ & \shortstack{\textbf{\textcolor{teal}{Impro-}}\\\textbf{\textcolor{teal}{vement}}} &
  \texttt{D}$_L$ & \texttt{D}$_{2L}^1$ & \texttt{D}$_{2L}^\infty$ & \texttt{B} & \texttt{C}$_{0.2}$ & \shortstack{\textbf{\textcolor{teal}{Impro-}}\\\textbf{\textcolor{teal}{vement}}} &
  \texttt{D}$_L$ & \texttt{D}$_{2L}^1$ & \texttt{D}$_{2L}^\infty$ & \texttt{B} & \texttt{C}$_{0.2}$ & \shortstack{\textbf{\textcolor{teal}{Impro-}}\\\textbf{\textcolor{teal}{vement}}} \\
\midrule
\multicolumn{25}{||c||}{\large \textbf{BLW-TrafficDS} ($\times\textbf{10}^\textbf{3}, L = 8$)} \\
\midrule
$\mathcal{I}_1$ & 167.1 & 172.3 & 164.9 & \textbf{163.8} & 218.0 & \textcolor{teal}{0.0\%} & 127.5 & 148.1 & 138.0 & \textbf{125.9} & 143.7 & \textcolor{teal}{0.0\%} & \underline{\textbf{102.8}} & 158.2 & 117.8 & 124.1 & 118.1 & \textcolor{teal}{17.2\%} & \textbf{119.3} & 121.1 & 122.8 & 130.8 & 133.1 & \textcolor{teal}{8.8\%} \\
$\mathcal{I}_2$ & 76.0 & \textbf{74.9} & 75.7 & 82.4 & 84.8 & \textcolor{teal}{9.1\%} & 40.1 & \textbf{37.8} & 41.7 & 83.0 & 50.7 & \textcolor{teal}{54.5\%} & 31.1 & \underline{\textbf{30.4}} & 35.3 & 77.9 & 34.6 & \textcolor{teal}{60.9\%} & \textbf{74.7} & 75.9 & 75.7 & 77.3 & 75.6 & \textcolor{teal}{3.4\%} \\
$\mathcal{I}_3$ & 56.7 & \textbf{56.0} & 56.7 & 62.0 & 58.8 & \textcolor{teal}{9.7\%} & 24.3 & \textbf{22.9} & 23.8 & 64.9 & 32.0 & \textcolor{teal}{64.7\%} & 18.3 & \underline{\textbf{17.2}} & 20.8 & 60.3 & 22.5 & \textcolor{teal}{71.5\%} & 57.1 & 58.4 & 57.6 & 59.5 & \textbf{56.0} & \textcolor{teal}{5.9\%} \\
$\mathcal{I}_4$ & 44.2 & 43.8 & 44.4 & 48.9 & \textbf{43.4} & \textcolor{teal}{11.2\%} & 16.1 & \textbf{15.4} & 15.8 & 48.1 & 20.8 & \textcolor{teal}{68.0\%} & 11.4 & \underline{\textbf{10.4}} & 12.6 & 47.2 & 15.2 & \textcolor{teal}{78.0\%} & 44.0 & 44.6 & 44.1 & 46.8 & \textbf{42.1} & \textcolor{teal}{10.0\%} \\
$\mathcal{I}_5$ & 33.5 & 33.4 & 33.8 & 37.6 & \textbf{31.7} & \textcolor{teal}{15.7\%} & 10.8 & \textbf{10.2} & 10.4 & 35.8 & 13.1 & \textcolor{teal}{71.5\%} & 7.1 & \underline{\textbf{6.5}} & 7.9 & 34.9 & 9.8 & \textcolor{teal}{81.4\%} & 31.4 & 32.1 & 31.9 & 35.5 & \textbf{28.7} & \textcolor{teal}{19.2\%} \\
$\mathcal{I}_6$ & 28.1 & 27.9 & 28.2 & 31.8 & \textbf{26.4} & \textcolor{teal}{16.9\%} & 8.0 & \textbf{7.8} & 8.0 & 30.2 & 9.3 & \textcolor{teal}{74.2\%} & 4.5 & \underline{\textbf{4.0}} & 5.0 & 29.4 & 6.8 & \textcolor{teal}{86.3\%} & 25.2 & 25.7 & 25.7 & 29.8 & \textbf{23.3} & \textcolor{teal}{21.8\%} \\
$\mathcal{I}_7$ & 19.3 & 19.2 & 19.4 & 21.7 & \textbf{18.8} & \textcolor{teal}{13.4\%} & \textbf{5.5} & 5.5 & 5.6 & 21.8 & 6.1 & \textcolor{teal}{74.8\%} & 2.7 & \underline{\textbf{2.4}} & 2.8 & 20.4 & 3.9 & \textcolor{teal}{88.2\%} & 19.2 & 19.5 & 19.5 & 20.6 & \textbf{18.1} & \textcolor{teal}{12.1\%} \\
$\mathcal{I}_8$ & 11.1 & 11.1 & 11.2 & 12.5 & \textbf{10.9} & \textcolor{teal}{12.8\%} & \textbf{3.3} & 3.5 & 3.4 & 12.1 & 3.6 & \textcolor{teal}{72.7\%} & 1.7 & \underline{\textbf{1.5}} & 1.8 & 11.7 & 2.6 & \textcolor{teal}{87.2\%} & \textbf{11.2} & 11.2 & 11.2 & 11.8 & 11.4 & \textcolor{teal}{5.1\%} \\
\midrule
$\mathbf{\overline{\mathcal{I}_1\text{--}\mathcal{I}_8}}$  \cellcolor{gray!0}  & \cellcolor{gray!0} \textbf{54.0} & \cellcolor{gray!0} 54.8 & \cellcolor{gray!0} 54.3 &  \cellcolor{gray!0}57.6 &  \cellcolor{gray!0}61.6 & \cellcolor{gray!0} \textcolor{teal}{6.3\%} & \cellcolor{gray!0} \textbf{29.4} &  \cellcolor{gray!0}31.4 &  \cellcolor{gray!0}30.9 & \cellcolor{gray!0} 52.7 & \cellcolor{gray!0} 34.9 & \cellcolor{gray!0} \textcolor{teal}{44.2\%} &  \cellcolor{gray!0}\underline{\textbf{22.4}} & \cellcolor{gray!0} 28.8 & \cellcolor{gray!0} 25.5 &  \cellcolor{gray!0}50.7 &  \cellcolor{gray!0}26.7 & \textcolor{teal}{55.8\%}  \cellcolor{gray!0}& \textbf{47.8} \cellcolor{gray!0} & 48.6  \cellcolor{gray!0}& 48.6  \cellcolor{gray!0}& 51.5  \cellcolor{gray!0}& 48.5  \cellcolor{gray!0}& \textcolor{teal}{7.2\%} \cellcolor{gray!0} \\
\midrule
\midrule
\multicolumn{25}{||c||}{\large\textbf{\texttt{Traffic} ($\times$500, $L = 4$)}} \\
\midrule
$\mathcal{I}_1$ & 1.70 & \textbf{1.53} & 1.67 & 1.79 & 1.82 & \textcolor{teal}{14.5\%} & 1.90 & \textbf{1.71} & 1.73 & 2.44 & 2.17 & \textcolor{teal}{29.9\%} & \underline{\textbf{1.08}} & 1.02 & 1.06 & 1.41 & 1.11 & \textcolor{teal}{27.7\%} & 1.29 & \textbf{1.30} & 1.32 & 1.41 & 1.36 & \textcolor{teal}{8.5\%} \\
$\mathcal{I}_2$ & 1.65 & \textbf{1.35} & 1.44 & 1.99 & 1.71 & \textcolor{teal}{32.2\%} & 1.25 & \textbf{1.23} & 1.27 & 2.02 & 1.52 & \textcolor{teal}{39.2\%} & 0.91 & 0.90 & 0.93 & 1.42 & \underline{\textbf{0.89}} & \textcolor{teal}{37.3\%} & \textbf{1.19} & 1.22 & 1.24 & 1.50 & 1.13 & \textcolor{teal}{20.7\%} \\
$\mathcal{I}_3$ & 0.61 & \textbf{0.44} & 0.47 & 0.81 & 0.62 & \textcolor{teal}{45.7\%} & 0.39 & \textbf{0.36} & 0.41 & 0.74 & 0.51 & \textcolor{teal}{51.5\%} & \underline{\textbf{0.32}} & 0.30 & 0.35 & 0.65 & 0.30 & \textcolor{teal}{53.9\%} & 0.53 & \textbf{0.54} & 0.55 & 0.66 & 0.53 & \textcolor{teal}{19.7\%} \\
$\mathcal{I}_4$ & 0.45 & \textbf{0.31} & 0.33 & 0.59 & 0.45 & \textcolor{teal}{47.5\%} & 0.22 & \textbf{0.20} & 0.24 & 0.59 & 0.35 & \textcolor{teal}{66.2\%} & 0.16 & \underline{\textbf{0.16}} & 0.19 & 0.53 & 0.16 & \textcolor{teal}{69.8\%} & 0.35 & \textbf{0.36} & 0.37 & 0.51 & 0.35 & \textcolor{teal}{31.4\%} \\
\midrule
\cellcolor{gray!0}$\mathbf{	\overline{\mathcal{I}_1\text{--}\mathcal{I}_4}}$ & \cellcolor{gray!0}1.10 & \cellcolor{gray!0}\textbf{0.90} & \cellcolor{gray!0}0.98 & \cellcolor{gray!0}1.29 & \cellcolor{gray!0}1.15 & \cellcolor{gray!0}\textcolor{teal}{30.2\%} & \cellcolor{gray!0}0.94 & \cellcolor{gray!0}\textbf{0.88} & \cellcolor{gray!0}0.91 & \cellcolor{gray!0}1.45 & \cellcolor{gray!0}1.14 & \cellcolor{gray!0}\textcolor{teal}{39.3\%} & \cellcolor{gray!0}0.62 & \cellcolor{gray!0}\underline{\textbf{0.59}} & \cellcolor{gray!0}0.63 & \cellcolor{gray!0}1.01 & \cellcolor{gray!0}0.62 & \cellcolor{gray!0}\textcolor{teal}{41.6\%} & \cellcolor{gray!0}\textbf{0.84} & \cellcolor{gray!0}0.85 & \cellcolor{gray!0}0.87 & \cellcolor{gray!0}1.02 & \cellcolor{gray!0}0.84 & \cellcolor{gray!0}\textcolor{teal}{17.6\%} \\
\midrule
\multicolumn{25}{||c||}{\large\textbf{\texttt{Weather}} ($\times\textbf{2}$, L = 4)} \\
\midrule
$\mathcal{I}_1$ & 1.05 & 0.76 & \textbf{0.75} & 0.78 & 1.32 & \textcolor{teal}{3.9\%} & 2.25 & \textbf{0.73} & 0.71 & 1.00 & 1.94 & \textcolor{teal}{27.0\%} & \underline{\textbf{0.40}} & 0.67 & 0.53 & 0.65 & 1.17 & \textcolor{teal}{38.5\%} & 0.49 & \textbf{0.48} & 0.49 & 0.68 & 1.72 & \textcolor{teal}{29.4\%} \\
$\mathcal{I}_2$ & 0.36 & \textbf{0.23} & 0.29 & 0.42 & 0.36 & \textcolor{teal}{45.2\%} & 0.37 & \textbf{0.22} & 0.26 & 0.54 & 0.43 & \textcolor{teal}{59.3\%} & \underline{\textbf{0.22}} & 0.23 & 0.28 & 0.37 & 0.22 & \textcolor{teal}{40.5\%} & \textbf{0.32} & 0.32 & 0.32 & 0.42 & 0.49 & \textcolor{teal}{23.8\%} \\
$\mathcal{I}_3$ & 0.13 & \textbf{0.09} & 0.10 & 0.20 & 0.26 & \textcolor{teal}{55.0\%} & 0.19 & \textbf{0.08} & 0.09 & 0.26 & 0.25 & \textcolor{teal}{69.2\%} & \underline{\textbf{0.08}} & 0.09 & 0.11 & 0.22 & 0.08 & \textcolor{teal}{63.6\%} & 0.18 & \textbf{0.17} & 0.18 & 0.23 & 0.39 & \textcolor{teal}{26.1\%} \\
$\mathcal{I}_4$ & 0.16 & \textbf{0.14} & 0.21 & 0.21 & 0.27 & \textcolor{teal}{33.3\%} & 0.17 & \textbf{0.12} & 0.13 & 0.27 & 0.26 & \textcolor{teal}{55.6\%} & \underline{\textbf{0.10}} & 0.13 & 0.14 & 0.23 & 0.11 & \textcolor{teal}{56.5\%} & 0.19 & \textbf{0.19} & 0.19 & 0.21 & 0.38 & \textcolor{teal}{9.5\%} \\
\midrule
\cellcolor{gray!0}$\mathbf{	\overline{\mathcal{I}_1\text{--}\mathcal{I}_4}}$ & \cellcolor{gray!0}0.43 & \cellcolor{gray!0}\textbf{0.31} & \cellcolor{gray!0}0.34 & \cellcolor{gray!0}0.40 & \cellcolor{gray!0}0.55 & \cellcolor{gray!0}\textcolor{teal}{22.5\%} & \cellcolor{gray!0}0.75 & \cellcolor{gray!0}\textbf{0.29} & \cellcolor{gray!0}0.30 & \cellcolor{gray!0}0.52 & \cellcolor{gray!0}0.72 & \cellcolor{gray!0}\textcolor{teal}{44.2\%} & \cellcolor{gray!0}\underline{\textbf{0.20}} & \cellcolor{gray!0}0.28 & \cellcolor{gray!0}0.26 & \cellcolor{gray!0}0.37 & \cellcolor{gray!0}0.40 & \cellcolor{gray!0}\textcolor{teal}{46.0\%} & \cellcolor{gray!0}\textbf{0.30} & \cellcolor{gray!0}0.29 & \cellcolor{gray!0}0.29 & \cellcolor{gray!0}0.38 & \cellcolor{gray!0}0.74 & \cellcolor{gray!0}\textcolor{teal}{23.7\%} \\

\midrule
\multicolumn{25}{||c||}{\large\textbf{\texttt{Electricity}  ($\times\mathbf{10}^{\mathbf{-1}}, L = 4$)}} \\
\midrule
$\mathcal{I}_1$ & 3.81 & 3.82 & 3.82 & \textbf{3.77} & 3.95 & \textcolor{teal}{0.0\%} & \textbf{3.83} & 3.95 & 3.95 & 4.50 & 4.86 & \textcolor{teal}{14.9\%} & 3.33 & 4.66 & 4.69 & \underline{\textbf{3.32}} & 3.78 & \textcolor{teal}{0.0\%} & 12.8 & 10.3 & 10.5 & \textbf{3.26} & 3.67 & \textcolor{teal}{0.0\%} \\
$\mathcal{I}_2$ & 1.15 & 1.16 & 1.16 & 1.17 & \textbf{1.14} & \textcolor{teal}{2.23\%} & 1.11 & \textbf{1.07} & 1.09 & 1.41 & 1.37 & \textcolor{teal}{24.3\%} & \underline{\textbf{0.919}} & 1.24 & 1.34 & 1.01 & 0.990 & \textcolor{teal}{9.37\%} & \textbf{4.67} & 3.69 & 3.84 & 1.05 & 1.02 & \textcolor{teal}{2.48\%} \\
$\mathcal{I}_3$ & 0.916 & 0.929 & 0.932 & 0.955 & \textbf{0.890} & \textcolor{teal}{6.81\%} & 0.807 & \textbf{0.782} & 0.836 & 1.08 & 0.984 & \textcolor{teal}{27.6\%} & \underline{\textbf{0.671}} & 0.852 & 0.931 & 0.835 & 0.689 & \textcolor{teal}{19.6\%} & 2.08 & 1.84 & 1.98 & 0.861 & \textbf{0.822} & \textcolor{teal}{4.53\%} \\
$\mathcal{I}_4$ & 0.369 & 0.379 & 0.380 & 0.399 & \textbf{0.353} & \textcolor{teal}{7.52\%} & 0.297 & \textbf{0.277} & 0.288 & 0.449 & 0.388 & \textcolor{teal}{38.3\%} & \underline{\textbf{0.238}} & 0.248 & 0.272 & 0.370 & 0.243 & \textcolor{teal}{35.7\%} & 0.567 & \textbf{0.509} & 0.522 & 0.372 & 0.322 & \textcolor{teal}{13.4\%} \\
\midrule
\cellcolor{gray!0}$\mathbf{	\overline{\mathcal{I}_1\text{--}\mathcal{I}_4}}$ & \cellcolor{gray!0}\textbf{1.56} & \cellcolor{gray!0}1.57 & \cellcolor{gray!0}1.57 & \cellcolor{gray!0}1.57 & \cellcolor{gray!0}1.58 & \cellcolor{gray!0}\textcolor{teal}{0.7\%} & \cellcolor{gray!0}\textbf{1.51} & \cellcolor{gray!0}1.52 & \cellcolor{gray!0}1.54 & \cellcolor{gray!0}1.86 & \cellcolor{gray!0}1.90 & \cellcolor{gray!0}\textcolor{teal}{18.8\%} & \cellcolor{gray!0}\underline{\textbf{1.29}} & \cellcolor{gray!0}1.75 & \cellcolor{gray!0}1.81 & \cellcolor{gray!0}1.39 & \cellcolor{gray!0}1.42 & \cellcolor{gray!0}\textcolor{teal}{6.93\%} & \cellcolor{gray!0}5.02 & \cellcolor{gray!0}4.09 & \cellcolor{gray!0}4.21 & \cellcolor{gray!0}\textbf{1.39} & \cellcolor{gray!0}1.46 & \cellcolor{gray!0}\textcolor{teal}{0.0\%} \\
\midrule
\multicolumn{25}{||c||}{\large\textbf{\synthds{}} ($\times\mathbf{10}^\mathbf{3}, L = 4$)} \\
\midrule
$\mathcal{I}_1$ & 15.0 & \textbf{14.5} & 18.5 & 50.9 & 24.8 & \textcolor{teal}{71.5\%} & 16.2 & \textbf{15.7} & 17.9 & 45.5 & 17.5 & \textcolor{teal}{65.5\%} & \underline{\textbf{13.2}} & 13.2 & 15.1 & 37.9 & 15.3 & \textcolor{teal}{65.2\%} & 14.1 & \textbf{14.1} & 17.5 & 40.2 & 20.0 & \textcolor{teal}{65.0\%} \\
$\mathcal{I}_2$ & \textbf{11.7} & 11.9 & 12.8 & 20.3 & 12.8 & \textcolor{teal}{42.4\%} & \textbf{11.4} & 12.2 & 15.0 & 18.8 & 13.5 & \textcolor{teal}{39.4\%} & \textbf{9.08} & 10.2 & 11.2 & 26.8 & 11.0 & \textcolor{teal}{66.2\%} & \underline{\textbf{7.96}} & 9.17 & 9.65 & 26.2 & 10.6 & \textcolor{teal}{69.6\%} \\
$\mathcal{I}_3$ & 11.7 & 11.2 & 10.7 & 17.1 & \textbf{10.6} & \textcolor{teal}{38.1\%} & 11.5 & 11.7 & 13.3 & 13.9 & \textbf{10.8} & \textcolor{teal}{22.4\%} & \underline{\textbf{7.74}} & 8.61 & 8.92 & 14.9 & 9.19 & \textcolor{teal}{48.0\%} & \textbf{7.97} & 8.38 & 8.63 & 16.2 & 8.71 & \textcolor{teal}{50.9\%} \\
$\mathcal{I}_4$ & 16.8 & \textbf{16.8} & 18.2 & 41.0 & 20.3 & \textcolor{teal}{59.1\%} & \textbf{15.5} & 17.3 & 19.7 & 34.9 & 16.6 & \textcolor{teal}{55.5\%} & \textbf{12.7} & 12.9 & 13.3 & 36.8 & 13.7 & \textcolor{teal}{65.6\%} & 13.1 & \textbf{12.4} & 13.6 & 33.7 & 13.4 & \textcolor{teal}{63.2\%} \\
\midrule
$\mathbf{	\overline{\mathcal{I}_1\text{--}\mathcal{I}_4}}$ & 13.8 & \textbf{13.6} & 15.1 & 32.3 & 17.1 & \textcolor{teal}{57.9\%} & \textbf{13.7} & 14.2 & 16.5 & 28.3 & 14.6 & \textcolor{teal}{51.7\%} & \underline{\textbf{10.7}} & 11.2 & 12.1 & 29.1 & 12.3 & \textcolor{teal}{63.3\%} & \textbf{10.8} & 11.0 & 12.3 & 29.1 & 13.2 & \textcolor{teal}{62.9\%} \\

\bottomrule

\end{tabular}
\end{adjustbox}
\caption{MAE values ($\downarrow$ is better) and improvement percentages ($\uparrow$ is better) relative to the baseline policy  are reported for four models: \texttt{DLinear}, \texttt{TimeMixer}, \texttt{iTransformer}, and \texttt{PatchTST}, trained using five distinct policies: \texttt{D$_L$-Policy}, \texttt{D$^1_{2L}$-Policy}, \texttt{D$^\infty_{2L}$-Policy}, \texttt{B-Policy}, and \texttt{C\textsubscript{0.2}-Policy}. The experiments use $L=8$ target intervals for the \wirelessds{} trace and $L=4$ for all other datasets. MAE values are scaled for readability: $\times 10^3$ for \synthds{} and \wirelessds{}, $\times 500$ for \texttt{Traffic}, $\times 2$ for \texttt{Weather}, and $\times 10^{-1}$ for \texttt{Electricity}. For \wirelessds{}, the number of intervals is increased to 8 and 16 for patching-based strategies. Bold values indicate the best MAE per model within each column, while underlined values denote the overall lowest MAE across all models. Model improvements are evaluated by comparing the best-performing policy to the baseline. The intervals of interest, denoted as $\mathcal{I}_1$--$\mathcal{I}_4$, correspond to a partition of the time series domain $\mathcal{X}$ into four equal-length contiguous sub-intervals. Similarly, $\mathcal{I}_1$--$\mathcal{I}_8$ indicates eight such sub-intervals. Notation $\overline{\mathcal{I}_1\text{--}\mathcal{I}_L}$ denotes the average performance over all intervals.
}
\label{tab:mae_transposed_all_with_improvement_col}
\end{table*}

\paragraph{Training Configuration.} To ensure the reproducibility of our experiments, we provide a comprehensive description of the training setup. The datasets were divided into training, validation, and testing subsets. A 66-17-17 split was used for the \synthds{}, \texttt{Traffic}, \texttt{Electricity}, and \texttt{Weather} datasets, while the \wirelessds{} dataset used a 70-10-20 split. Four state-of-the-art time series forecasting models---iTransformer, DLinear, PatchTST, and TimeMixer---were evaluated across the designated tasks.

The \wirelessds{}, \texttt{Traffic}, \texttt{Electricity}, and \texttt{Weather} datasets were processed using a multivariate-to-multivariate configuration, whereas the \synthds{} dataset used a univariate-to-univariate setup. Input sequence lengths and forecasting horizons were tailored to each dataset: 48/24 for \synthds{}, 96/24 for \wirelessds{}, and 168/48 for \texttt{Traffic}, \texttt{Electricity}, and \texttt{Weather}. The number of input channels was cropped to 100 for all datasets except \texttt{Weather}, which used 21 channels due to its limited dimensionality.

Model configurations across all experiments included 3 layers with a model dimension of 256. TimeMixer was configured with a channel dimension of 100. All models were trained with a batch size of 32 for 50 epochs. The AdamW optimizer~\cite{loshchilov2017decoupled} was used with an initial learning rate of $10^{-3}$ and a cosine annealing schedule with $\eta\_{\min} = 10^{-5}$, defined as:$\eta_{t} = \eta_{\min} + 0.5 (\eta_{\max} - \eta_{\min}) (1 + \cos(\pi \cdot t / n_{\text{epochs}}))$.

The MAE loss was used for regression tasks, while Binary Cross Entropy loss was applied for classification tasks when applicable. Early stopping with a patience of 5 epochs and checkpoint saving mechanisms were employed to reduce overfitting. Validation loss was computed as the average loss over each training interval.

The \discretePacthedPolicy{} was evaluated using 4 and 8 intervals for the \synthds{}, \texttt{Traffic}, \texttt{Electricity}, and \texttt{Weather} datasets, and 8 and 16 intervals for the \wirelessds{} dataset. Interval endpoint sampling was performed per sample within each batch and incorporated into the forward pass as detailed in Section~\ref{sec:methodology}.

The dataset sizes are as follows: \synthds{}---3,456 points, \wirelessds{}---1,025 points,  \texttt{Weather}---52,696 points, \texttt{Traffic}---17,544 points, and \texttt{Electricity}---26,304 points. To set forecasting boundaries, the maximum values were assigned as $\max\parentheses{\mathcal X}\in \set{0.2, 500.0, 10000.0}$ for \texttt{Traffic}, \texttt{Weather}, and \texttt{Electricity}, respectively. All experiments were executed on a system with 6 Tesla V100/16GB GPUs and an Intel(R) Xeon(R) Platinum 8164 CPU (104 cores).

\subsection{Numerical Results} \label{subsec:nrres}

\paragraph{Qualitative Comparison of Training Schemes.} To build intuition about how different training policies behave when downstream tasks specify an interval of interest, thereby prioritizing accuracy within that interval, we begin by evaluating the iTransformer model on the \texttt{SynthDS} trace. This dataset is chosen for its simplicity, which facilitates both result interpretation and qualitative performance assessment.

\emph{Baseline.} We first examine the averaging behavior of the baseline policy when training iTransformer on \texttt{SynthDS}. As shown in Figure~\ref{fig:avgtend}, the model converges to an averaged prediction centered at the trace’s midpoint~($\nicefrac{1}{2}$), reflecting its insensitivity to specific intervals of interest. This is undesirable in scenarios where the conditional mean within the interval of interest differs from the global mean as the model fails to adapt its predictions to the interval-specific distribution.

\emph{End-to-End Training Approach.} Figure~\ref{fig:e2e_unif_a} shows the performance of \taskSpecificPolicy{} with the iTransformer model, targeting the specific interval $\mathcal{I} = [0.75, 1]$. This corresponds to an end-to-end strategy in which the forecasting loss is adapted to emphasize task-relevant intervals. The selected interval represents a distinct hypothesis, and this policy serves as an optimal benchmark, illustrating the best-case performance achievable by a method that can adapt to any interval and be configured at inference time.

\emph{Continuous Exploration of Intervals.} We also evaluate \uniformPolicy{} using the iTransformer model on \texttt{SynthDS}. The results in Figure~\ref{fig:e2e_unif_b} show that this approach struggles to effectively learn the relationship between intervals of interest and their corresponding hypotheses, underscoring the difficulty of the learning task even in a synthetic setting. In this configuration, the model is tasked with learning a mapping from every possible interval $\mathcal{I} \subset \mathcal{X}$ to its associated hypothesis. We further explore, in Figure~\ref{fig:uniform_delta}, the impact of restricting the support space of training intervals by enforcing a minimum separation $|\mathcal{I}| \geq \delta$. While performance improves with a reduced sampling space, beyond a certain threshold $\delta'$, it deteriorates as the model overshoots the true interval lengths, introducing bias.

\emph{Targeted Exploration of Intervals.} Building on the observed limitations of continuous exploration, Figure \ref{fig:disc_patch_4} examines a discrete policy (\discretePolicy{}) on \texttt{SynthDS}. This policy uniformly samples from a predefined, finite, and relatively small set of intervals of interest during training. We first evaluate it using the optimal \taskSpecificPolicy{}, which can  distinguish signal patterns. The results show that \discretePolicy{} achieves performance comparable to \taskSpecificPolicy{} while retaining the advantage of inference-time adaptability to diverse downstream tasks. However, this approach assumes prior knowledge of the relevant interval set (or its size), motivating the subsequent experiment in which we intentionally increase the number of considered intervals to assess the reconstructability of the true signal.

\emph{Adaptive Interval Exploration with Patching Techniques.} Figures \ref{fig:disc_patch_8} and \ref{fig:disc_patch_8_1} evaluate the effectiveness of our patching strategy, described in Section \ref{sec:methodology}, in capturing the underlying dynamics of \texttt{SynthDS}. These experiments consider scenarios in which the number of intervals is deliberately overestimated during training and subsequently patched back to an unknown target interval at inference time. This overestimation is intended to qualitatively assess the robustness of the patching schemes. While the schemes exhibit visually similar performance, a quantitative analysis is required for further comparison. Notably, both approaches distinguish the four distinct underlying hypotheses used to construct the synthetic trace compared to \uniformPolicy{}, thereby motivating our design choice: train a model on a large set of intervals that can be flexibly combined at inference time to produce focused forecasts for a specific interval of interest.

\paragraph{Quantitative Comparison of Training Schemes.} Following our qualitative evaluation of the proposed training policies, Table~\ref{tab:mae_transposed_all_with_improvement_col} presents the numerical results obtained on the \wirelessds{}, \texttt{Traffic}, \texttt{Weather}, \texttt{Electricity}, and \synthds{} datasets. The table reports the MAE values for the four models, each trained under the four distinct policies.

Table~\ref{tab:mae_transposed_all_with_improvement_col} reports the results of training the different models on the \synthds{} trace. We observe comparable performance between the iTransformer and PatchTST models. Notably, \baselinePolicy{}, which does not support inference-time adaptation of the interval of interest, yields the poorest performance. The \uniformPolicy{} shows a modest MAE improvement, suggesting that training toward specific intervals provides some benefit. However, sampling all possible intervals in \uniformPolicy{} hinders the model’s ability to learn distinct underlying hypotheses, as evidenced by the superior performance of \discretePolicy{}, configured here with four specific intervals. Introducing the patching mechanism in \discretePacthedPolicy{} leads to some performance degradation for certain intervals; nonetheless, its overall MAE remains comparable to, and occasionally exceeds, that of \discretePolicy{}, potentially due to an ensemble-like noise reduction effect.  

These conclusions generally extend to the real-world traces. Interestingly, in some cases patching outperforms the vanilla discrete approach, with averaging-based patching achieving the best performance for the PatchTST model. Results differ for weaker models such as DLinear, where \uniformPolicy{} appears to dominate; however, given its substantially lower overall accuracy compared to the other models, the ordering of improvements may be influenced by random factors. We also note that \discretePolicy{} is optimally configured here with intervals that precisely align with the true underlying hypotheses under the \synthds{} trace---a level of prior knowledge unlikely in practical applications. Consequently, \discretePacthedPolicy{} offers a more pragmatic alternative, providing a viable strategy for training foundation models capable of adapting to varying intervals during inference.
\subsection{Additional Numerical Experiments}
\subsubsection{Implications for Downstream Task: Forecasting for Energy-Saving}
We consider a heterogeneous wireless network comprising two tiers: a \emph{capacity cell} (e.g., a small or micro cell) providing high-throughput service within a localized area, and an overlaid \emph{coverage cell} (e.g., a macro cell) ensuring broad-area connectivity. Two-tier architectures with dynamic small cell control have emerged as a key strategy for balancing spectral efficiency and energy savings in 5G and beyond \cite{Celebi2019LoadOnOff, Ju2022EnergyEffUDN, Wu2021DeepBSC, SalvatLozano2025Kairos, 3GPP2014TR36927}. In particular, recent work has shown that monitoring DLPRB utilization provides an effective signal for real-time sleep mode activation in energy-aware networks.  We assume that the capacity cell can be selectively deactivated to reduce energy consumption when traffic demand is low. To govern this behavior, we adopt a threshold-based energy-saving policy that operates based on the observed {DLPRB} utilization. This mirrors widely-used control strategies that compare traffic load against fixed or learned thresholds to trigger cell activation or sleep~\cite{Celebi2019LoadOnOff, Ju2022EnergyEffUDN}.

Let $ u(t) \in [0,1] $ denote the normalized DLPRB utilization at discrete time index $ t \in \mathbb{Z}_{\geq 0} $, and let $ u_{\mathrm{th}} \in [0,1] $ represent a fixed utilization threshold. The binary state variable $ S(t) \in \{0,1\} $ indicates whether the capacity cell is active ($ S(t) = 1 $) or deactivated ($ S(t) = 0 $). The policy is defined as:
\begin{align}
S(t) =
\begin{cases}
1, & \text{if } u(t) \geq u_{\mathrm{th}}, \\
0, & \text{otherwise}.
\end{cases}
\end{align}
The instantaneous cell traffic load, denoted $ L(t) $, is computed as the product of DLPRB utilization and the maximum throughput capacity of the capacity cell:
\begin{align}
L(t) \triangleq u(t) \cdot C_{\mathrm{cap}},
\end{align}
where $ C_{\mathrm{cap}} > 0 $ is the peak service rate (in Mbps) of the capacity cell.

\paragraph{Realized Throughput.} The realized throughput, $ R(t) $, is constrained by the physical limits of the serving infrastructure. When the capacity cell is active, it serves the traffic directly, up to its maximum capacity. If deactivated, the traffic is offloaded to the coverage cell, subject to its own capacity constraint and a possible degradation in service quality. This is formalized as:
\begin{align}
R(t) =
\begin{cases}
\min\left( L(t), C_{\mathrm{cap}} \right), & \text{if } S(t) = 1, \\
\alpha \cdot \min\left( L(t), C_{\mathrm{cov}} \right), & \text{if } S(t) = 0,
\end{cases}
\end{align}
where $ C_{\mathrm{cov}} > 0 $ denotes the capacity of the coverage cell and $ \alpha \in (0,1] $ is a degradation factor that accounts for reduced performance when traffic is offloaded.

\paragraph{Energy Consumption.} To evaluate the energy-performance trade-off, we associate an energy cost with each state. Let $ E_{\mathrm{on}} $ and $ E_{\mathrm{off}} $ represent the per-time-unit energy consumption when the capacity cell is active or inactive, respectively. The total energy consumption at time $ t $ is then:
\begin{align}
E(t) = S(t) \cdot E_{\mathrm{on}} + (1 - S(t)) \cdot E_{\mathrm{off}}.
\end{align}

Over a time horizon of $ T $ intervals, the average throughput and average energy consumption are given by:
\begin{align}
\bar{R} = \frac{1}{T} \sum_{t=1}^{T} R(t), \quad \bar{E} = \frac{1}{T} \sum_{t=1}^{T} E(t).
\end{align}
This model captures the core dynamics of a load-aware energy-saving policy in a two-tier network, balancing energy efficiency with service quality under realistic physical constraints.
\paragraph{Downstream Application Optimization Problem.} We consider the following optimization problem that balances average throughput and energy consumption:
\begin{align}
\max_{u_{\mathrm{th}} \in [0,1]} \quad & (1 - \lambda)\, \bar{R}(u_{\mathrm{th}}) - \lambda\, \bar{E}(u_{\mathrm{th}}) \label{eq:objective}
\end{align}
where $ \lambda \in [0,1] $ is a scalar trade-off parameter that governs the relative importance of energy efficiency versus throughput performance. Let $ u^\star_{\mathrm{th}}(\lambda) $ denote the optimal threshold that solves problem~\eqref{eq:objective} for a given value of $ \lambda $. 

\paragraph{Instantiation and Numerical Results.} We instantiate the problem using representative parameter values from prior work~\cite{Celebi2019LoadOnOff, Ju2022EnergyEffUDN}. In particular, the capacity cell peak throughput was set to $C_{\mathrm{cap}} = 100$~Mbps, while the coverage cell peak throughput was $C_{\mathrm{cov}} = 30$~Mbps. An offloading degradation factor of $\alpha = 0.5$ was applied to model performance reduction during traffic redirection. The energy consumption of a capacity cell was configured as $E_{\mathrm{on}} = 1266$~Wh when active and $E_{\mathrm{off}} = 320$~Wh when inactive. These parameters define the operational conditions under which forecasting models were evaluated. 

Specifically, we train the {iTransformer} model on the \wirelessds{} dataset under two distinct policies: the \baselinePolicy{} and the \discretePacthedPolicy{} ($L = 4$ intervals), with a forecasting horizon length of $H = 24$, with a test set rolled over $96$ hours. We then evaluate and compare the performance of these policies. Since the threshold of interest is not known \textit{a priori}, we deliberately select a slightly broader interval, $[0, 0.5]$, to ensure coverage; this involves patching together two intervals for the \discretePacthedPolicy{}. Based on the generated forecasts, we fix a threshold value and observe the resulting sequence of decisions, denoted as $S(t)$ for $t \in [T]$, where $T$ is the length of the forecast horizon. We analyze the discrepancies in sleep durations induced by the forecasts under the different policies and restrict our attention to a specific range of thresholds: $U_{\mathrm{th}} \in [0, 0.025]$.

\begin{figure}[t!]
\centering
\includegraphics[width=.7\linewidth]{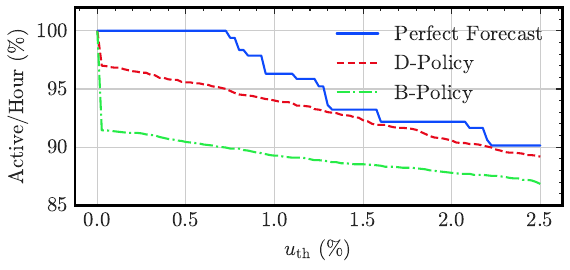}

\caption{The impact of forecasting accuracy on wireless network energy saving as downstream application.\label{fig:threholds}}
\end{figure}

As shown in Figure~\ref{fig:threholds}, the task-specific policy corresponding to a selected threshold yields decisions that are more closely aligned with the optimal policy---that is, the policy a decision-maker would follow with perfect foresight of future DLPRB values. This alignment demonstrates that the task-specific forecasts lead to better downstream optimization performance. Quantitatively, the task-specific policy produces forecasts that reduce the sleep duration error by a factor of three ($\times 3$), corresponding to an average reduction of one hour in sleep duration error per day. This improvement translates into an energy saving error of only $337$ watts, in contrast to a significantly higher mismatch of $0.950 $ kilowatts observed under the baseline policy per day.


\begin{figure}[t!]
    \centering
    \includegraphics[width=0.7\linewidth]{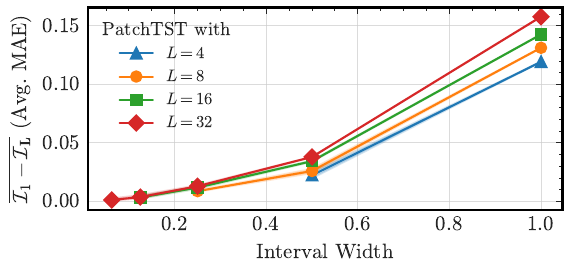}
    \caption{Comparison of the performance of the $\texttt D_4, \texttt D_8, \texttt D_{16}$ and $\texttt  D_{32}$ policies with a fixed decay rate for the PatchTST model. The MAE is averaged across the different intervals. }
    \label{fig:maecompdecay147}
\end{figure}

\begin{figure}[t!]
    \centering
    \includegraphics[width=0.7\linewidth]{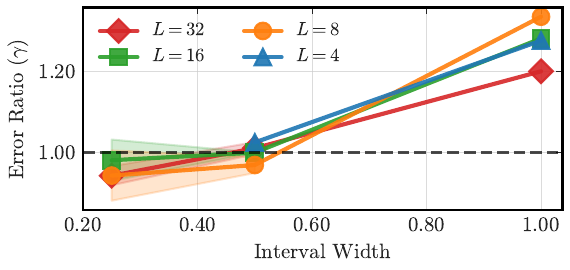}
    \caption{Comparison of the error ratio $\gamma$ across discretizations. Values $\gamma<1$ indicate that the $\infty$-strategy performs better, while $\gamma>1$ favors the $1$-strategy.}
    \label{fig:maecomparison1infstrat}
\end{figure}

\begin{figure}[t!]
    \centering
    \includegraphics[width=0.7\linewidth]{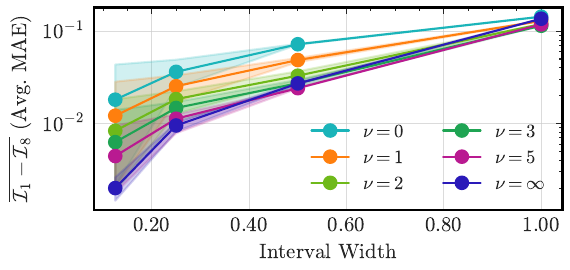}
    \caption{Comparison of the performance of the different decay rates for the PatchTST model trained with the $\texttt D_{8}$-Policy.}
    \label{fig:maecompint4}
\end{figure}

\begin{figure}[t!]
    \centering
    \includegraphics[width=0.7\linewidth]{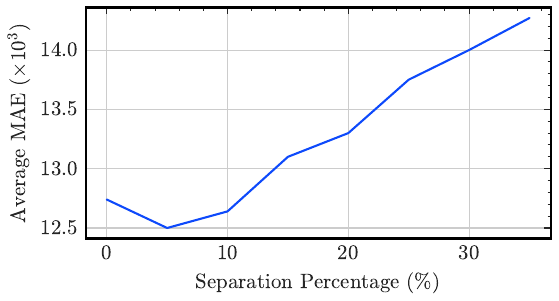}
    \caption{Interval-separation percentage ($\delta$)
impact on \texttt{C-Policy}.}
    \label{fig:uniform_delta}
\end{figure}

\subsubsection{Hyperparameter Sensitivity Analysis}
We conducted experiments on the specific hyperparameters introduced by our work, namely the number of intervals $L$ in which we divided the time series domain $\mathcal{X}$, the decay rate $\nu \in [0, \infty]$ and the  the patching strategies ($1$-strategy and $\infty$-strategy). 
\paragraph{Interval Granularity ($L$).} Among the hyperparameters introduced, the number of intervals $L$ plays the most critical role. It determines the granularity of the discretization over the time series domain $\mathcal{X}$, directly influencing both model complexity and performance. Figure~\ref{fig:maecompdecay147} illustrates the effect of varying $L$ on model performance. To assess this, we trained the PatchTST model on the \synthds{}, under four instances of \discretePacthedPolicy{}{}. Each policy corresponds to a partitioning of the domain $\mathcal{X}$ into $L \in \{4, 8, 16, 32\}$ intervals, with a fixed decay rate $\nu$ across all runs. For evaluation, we measured the MAE over coarser partitions obtained via powers-of-two binning.
The results demonstrate a clear trade-off: increasing the number of intervals improves granularity but may lead to decreased accuracy at coarser levels due to over-patching. This underscores the importance of selecting an appropriate value of $L$ to balance detail sensitivity with robustness across scales. For practical applications, a preliminary sweep over $L$ is recommended to identify the configuration that best controls patching frequency while maintaining predictive accuracy.
\paragraph{Patching Strategies Comparison ($\star$).}
To systematically investigate the performance trade-off between the $1$-strategy and the $\infty$-strategy, we formulated an evaluation strategy centered on the MAE across varying interval lengths. For this purpose, the PatchTST model, one of the best models from our empirical studies, was trained on the noiseless SynthDS dataset under four distinct policies: $\texttt  D_4$, $\texttt D_8$, $\texttt D_{16}$, and $\texttt D_{32}$. A comparative evaluation of the $1$-strategy and $\infty$-strategy was then conducted for each policy. This evaluation was performed on partitions of the codomain of the time-series where the total number of intervals corresponded to a power of two. Figure~\ref{fig:maecomparison1infstrat} presents a comparative analysis of the two patching mechanisms where we divided the MAE of the $\infty$-strategy by the MAE of the $1$-strategy and we observed the results over different intervals length.

We observe that for smaller intervals where the hypotheses are well-defined and there is no need for averaging, the $\infty$-strategy outperforms the $1$-strategy. As the intervals are getting larger, the need for averaging grows, thus making the $1$-strategy better. 

This analysis shows that for a dataset, when performing inference on small intervals or intervals with clearly defined hypotheses, the $\infty$-strategy is needed, while for larger intervals on which the B-Policy would average, the $1$-strategy is needed.

\paragraph{Decay Rate Analysis ($\nu$).}
In the same context as the experiment involving the number of intervals, we conducted an experiment related to the behaviors of the decay rate. We employed the $\texttt D_{32}$ policy and studied decay rates in the set $\{0, 1, 2, 5, \infty\}.$ The results of the experiment can be seen in Figure~\ref{fig:maecompint4}.

We observe that the decay rate $\nu$ plays a significant role in enhancing forecasting accuracy, particularly in scenarios where substantial patching is required, i.e., when the interval length is large. In the extreme case where all $32$ training intervals are aggregated into a single target interval (resulting in an interval length of $1$), the best performance is achieved when $\nu = \infty$, corresponding to a no-decay setting. However, as the intervals of interest become smaller, corresponding to finer-grained target partitions, we observe improved performance when a finite decay rate is applied. In these cases, the decay mechanism effectively introduces a soft overlap between training intervals, promoting smoother generalization across neighboring regions of the domain. This highlights the importance of tuning $\nu$ based on the level of aggregation or granularity used in the target interval structure.

\paragraph{Interval Separation ($\delta$).} As introduced in Section~\ref{sec:methodology}, the \uniformPolicy{} incorporates a hyperparameter, the interval separation $\delta$. In this section, we investigate its effect using the iTransformer model trained on the \synthds{} trace. Specifically, we vary $\delta$ within the range $[0, 0.4]$ over the domain $\mathcal{X} = [0,1]$.

Figure~\ref{fig:uniform_delta} illustrates the effect of constraining the sampling space of training intervals by enforcing a minimum separation constraint $\mathcal{I} \geq \delta$. Reducing the sampling space initially yields improved performance; however, beyond a critical threshold $\delta^\prime\approx 5\%$, performance degrades as the model tends to overestimate true interval lengths, thereby introducing bias. It is worth noting that in \synthds{}, the ground-truth hypotheses are separated by intervals of length $1/4$.

\section{Conclusion}\label{s:conclusion}
This research introduces a novel training methodology designed to transform existing TSF models into foundational architectures capable of inference-time adaptation for diverse downstream tasks. A comprehensive empirical evaluation was performed to ascertain the efficacy of the proposed approach. 

Future research should explore scaling the proposed models to diverse domains and fine-tuning existing time series foundation models (e.g., Timer~\cite{liu2024timer},  Chronos~\cite{ansari2024chronos}) to enable adaptation to arbitrary intervals, as presented herein. Furthermore, a theoretical analysis, potentially leveraging PAC-learning frameworks~\cite{shalev2014understanding}, to justify the performance disparity between the \uniformPolicy{} and \discretePacthedPolicy{} offers a promising direction for future investigation and potential framework enhancements.

\section{Acknowledgments} We are grateful to Prof. Zhi-Quan (Tom) Luo for his insightful and constructive feedback, which greatly advanced the development of this study. We also thank Dr. Dachao Lin and Dr. Xi Zheng for their valuable assistance in providing and curating the wireless dataset used in this research.

\bibliographystyle{plain}
\bibliography{main}

\end{document}

%% file: notation_defs.tex
\usepackage{blindtext}
\usepackage{enumitem}
\usepackage{xpatch}
\usepackage{bm}
\usepackage[framemethod=TikZ]{mdframed}
\usepackage{subcaption}
\usepackage{caption}
\usepackage{colortbl} 
\usepackage{stackengine}
\renewcommand{\vec}[1]{\bm{\mathbf{#1}}}

\newcommand{\w}{\vec{w}}

\newcommand{\interval}[1]{\left[#1\right]}
\newcommand\norm[1]{\lVert#1\rVert}

\newcommand{\reals}{\mathbb{R}}
\newcommand{\naturals}{\mathbb{N}}

\newcommand{\set}[1]{\left\{#1\right\}}
\newcommand{\parentheses}[1]{\left(#1\right)}

\newcommand{\card}[1]{\left|#1\right|}

\newcommand{\supp}[1]{\mathrm{supp}\parentheses{#1}}

\usepackage{dsfont}

\renewcommand{\w}{\vec{\w}}

\newcommand{\argmax}{\mathrm{argmax}}

\makeatletter
\DeclareMathOperator*{\argmin}{\smash[b]{\operator@font arg\,min}}
\makeatother
\newcommand{\baselinePolicy}[0]{\texttt{B-Policy}}
\newcommand{\taskSpecificPolicy}[0]{\texttt{E2E-Policy}}
\newcommand{\uniformPolicy}[0]{\texttt{C-Policy}}
\newcommand{\discretePolicy}[0]{\texttt{D$_L$-Policy}}
\newcommand{\discretePacthedPolicy}[0]{\texttt{D$_L^\star$-Policy}}
\usepackage{subcaption}
\usepackage{caption}
\newcommand{\loss}{L}
\usepackage{sidecap}
\newcommand{\wirelessds}{\texttt{BLW-TrafficDS}}
\newcommand{\synthds}{\texttt{SynthDS}}
\usepackage{longtable}
\usepackage{adjustbox}